\documentclass{SLT2024}
\usepackage{csquotes}

\interspeechcameraready
\title{PromptKWS: A Novel Prompt-Guided Open-Vocabulary Keyword Spotting Framework}
\name[affiliation={1}]{Gaopeng}{Xu}
\name[affiliation={2}]{Chengfei}{Li}
\name[affiliation={1}]{Xianliang}{Wang}

\name[affiliation={1}]{Lin}{Zhu}
\name[affiliation={1}]{Juan}{Wei}
\name[affiliation={1}]{Wenpeng}{Li}
\name[affiliation={1}]{Jianwei}{Niu}
\name[affiliation={1}]{Jie}{Gao}
\address{
  $^1$NIO, China \\
  $^2$Qilu Normal University, China 
  }
\email{gaopeng.xu@nio.com}


\begin{document}
%\linenumbers

\maketitle

\begin{abstract}
In this paper, we present PromptKWS, a novel Prompt-guided keyword spotting (KWS) framework to improve the accuracy of open vocabulary KWS systems. In specific terms, we introduce the Prompt Phrases Prediction Network (PPN), an encoder-decoder architecture designed to effectively extract keyword prompts embeddings. we employ the PPN encoder to encode the keyword prompts and infuse the prompt embedding into the Prompt-guided KWS encoder by utilizing a Prompt-acoustic Multi-head Cross-attention (MHCA). Experiments show that PromptKWS improves the wakeup rate by over 10\% compared to baseline system. Notably, another strength of PromptKWS is its ability to effectively leverage keyword prompts for adapting to complex real-world environments involving noise and pronunciation variations. In comparison to purely acoustic models, which often struggle in such situations, PromptKWS demonstrates remarkable performance, with an average accuracy improvement of over 15\% in test sets.
    
\end{abstract}

\section{Introduction}
Keyword spotting is a crucial component of many speech-based interaction systems, enabling the identification of specific words or phrases within continuous audio streams. KWS has numerous applications, including smart home devices and intelligent cockpits. While existing KWS models \cite{k1,k3,k4,k5,k6} can achieve high detection rates, developing user-defined or customizable systems remains a significant challenge, especially within the context of continuous speech. This is due, in part, to the need for a large dataset containing target keywords and the inflexibility of changing target keywords, which can hinder the expansion of KWS models to various applications.

Open-vocabulary KWS systems commonly utilize a two-stage approaches \cite{8462227}. These approaches typically involve an initial stage of acoustic modeling, succeeded by a keyword search stage. Keyword search, resembling the graph search process employed in automatic speech recognition (ASR), is commonly utilized for open-vocabulary KWS in continuous speech due to the relative ease of extending well-trained ASR systems to KWS tasks. In this context, customizable keywords are represented as brief sequences of acoustic modeling units, such as mono phones or graphemes, with the option of introducing a filler model to absorb non-keyword elements. Weighted Finite State Transformer (WFST) has been widely recognized as an efficient graph search method. When constructing a decoding graph \cite{8268943,pan,sun2017}, the key to this type of system is how to effectively design based on the scores of keywords and fill paths. With the rise of deep learning, \cite{chen2014} introduced acoustic models based on deep learning for keyword recognition (KWS) and equipped them with post-processing modules, thereby simplifying the training and inference process of the model. Subsequently, numerous studies \cite{shan2018,yang2022,9413588} have focused on further optimizing the acoustic model architecture to enhance the overall performance of the system.

Recent research has made significant advancements in text and image generation by using a text description as a prompt \cite{l1,l2,l3,l4,l5}. This approach has proven successful not only in generating text and images but also in other areas such as text to speech (TTS) and automatic speech recognition (ASR). For example, PromptTTS\cite{tts} can synthesize speech based on textual descriptions, enabling tasks such as prompted text-to-speech or text-to-audio. In the ASR domain, PromptASR\cite{asr} leverages natural language to guide the stylistic transformation in speech conversion. Inspired by PromptASR, we propose PromptKWS, a method that utilizes keyword prompt to improve the performance of open-vocabulary KWS. Unlike PromptASR, which uses a BERT model to extract prompt embedding, KWS model are often used on edge devices, which means that the overall parameter size needs to be much smaller than that of ASR models. Taking into account both performance and model size, we propose a new method called the PPN, which is a lightweight and efficient encoder-decoder architecture designed to extract embeddings from keyword prompts. Since the PPN decoder is only used during the training phase, we use only the PPN encoder for encoding during inference, which adds very few parameters to the model. Finally, we integrate the keyword prompt embedding into conformer-like Prompt-guided KWS encoder by utilizing the Prompt-acoustic MHCA and demonstrate the effectiveness of PromptKWS for open-vocabulary KWS tasks. We evaluate the proposed promptKWS framework on multiple test sets. Our experimental results demonstrate an impressively high wake-up rate using the proposed approach.

\begin{figure}[t]
  \centering
  \includegraphics[scale=0.6]{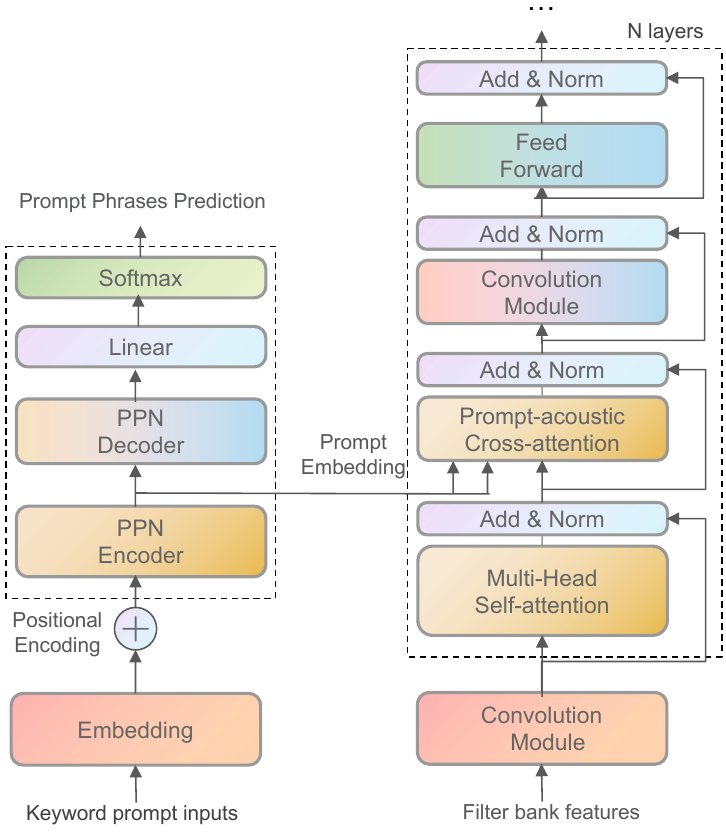}
  \caption{The architecture of PromptKWS.}
  \label{fig:promptkws}
\end{figure}

\section{PromptKWS}

\subsection{Model Architecture}

As depicted in Figure~\ref{fig:promptkws}, the proposed PromptKWS is composed of three main parts: a PPN module, a Prompt-guided KWS encoder (EnK), and a KWS decoder (DecK). The PPN is tasked with processing keyword prompts and generating prompt embedding. EnK consists of of N Conformer-like layers. This structure allows each layer within the encoder to receive not only acoustic feature but also keyword prompt embedding encoded by PPN. The integration of keyword prompt and speech modalities within EnK is achieved through cross-attention mechanisms. In this context, the keyword prompt embedding serve as key-value pairs, while the acoustic hidden states function as queries, facilitating a seamless blend of information from both modalities. The DecK incorporates a linear layer equipped with softmax activation, which transforms the Encoder's output into the ultimate probability distribution output.

\subsection{Prompt Phrases Prediction Network}

The Prompt Phrases Prediction Network (PPN) is a lightweight and efficient encoder-decoder architecture designed for keyword prompt embedding extraction. The encoder is responsible for encoding the input prompt phrase into a fixed-length vector, while the decoder generates a sequence of output tokens that represent the prompt phrase. By utilizing the PPN decoder output, we can compute the loss associated with prompt phrases, which enhances the discriminative ability of the keyword prompt embedding. Specifically, the decoder output is compared with the ground truth prompt phrase using a cross-entropy loss function. This loss encourages the model to generate output tokens that are consistent with the ground truth prompt phrase, thereby improving the quality of the embeddings extracted from the prompt phrases. One of the key advantages of the PPN is its lightweight architecture, which allows it to be easily integrated into existing KWS models without significantly increasing the model's parameters or computational complexity. Since the PPN decoder is only used during the training phase, we use only the PPN encoder for encoding during inference, which adds very few parameters to the model.

\subsection{Prompt-guided KWS Encoder}

The Prompt-guided KWS encoder component of the model is composed of N Conformer-like layers, each of which is carefully designed to include several key components, including FF layers, Masked MHSA layers, MHCA layers, Causal Convolution layers, and an additional set of FF layers.
The MHSA layer in the Encoder is responsible for processing the input speech data and generating a sequence of hidden states that capture the relevant acoustic features. The prompt contextual information E\textsubscript{prompt} is incorporated into the KWS encoder through the use of PPN module, which is responsible for encoding the keyword prompt into prompt embedding.
The integration of the prompt text and speech modalities is achieved through the use of Prompt-acoustic MHCA mechanisms, which allow each layer within the encoder to receive not only acoustic feature, but also prompt embedding encoded by PPN module. In this context, the prompt embedding serve as key-value pairs, while the acoustic hidden states MHSA\textsubscript{out} function as queries. The MHCA facilitates a seamless blend of information from both modalities, allowing the model to leverage contextual information from the keyword prompt to guide the keyword detection process. The entire operation is described in Equation~\ref{equation:eq1}.

\begin{equation}
\begin{aligned}
\mathbf{k} &= (\mathbf{E}_{prompt}) \mathbf{W}_{k}, \\  
\mathbf{v} &= (\mathbf{E}_{prompt}) \mathbf{W}_{v}, \\  
\mathbf{q} &= \left(\mathbf{MHSA}_{\text{out}}\right) \mathbf{W}_{q}, \\  
\mathbf{MHCA}_{\text{out}} &= \mathbf{MHCA}(\mathbf{q}, \mathbf{k}, \mathbf{v}), \\  
\mathbf{CA}_{\text{out}} &= \mathbf{MHSA}_{\text{out}} + \mathbf{MHCA}_{\text{out}}, \\  
\mathbf{y} &= \mathbf{CA}_{\text{out}} + \operatorname{Conv}(\mathbf{CA}_{\text{out}})
\end{aligned}
\label{equation:eq1}
\end{equation}

\subsection{KWS Decoder}
The KWS decoder component is implemented using Connectionist Temporal Classification (CTC) \cite{ctc}. The CTC decoder is responsible for generating a sequence of output tokens that represent the detected keyword. The CTC decoder consists of a linear layer and a softmax layer, which maps the hidden states from the encoder to a sequence of probability distributions over the output tokens.

\subsection{Details of Prompts}
The PromptKWS model includes two types of prompts: keyword prompt and scene prompt. The keyword prompt provides specific information about the wake word or trigger phrase that the model is trained to detect, including the corresponding text information of the wake word or trigger phrase. It also includes a special identifier "\textless no\_prompt\textgreater" to ensure that the model is not affected in the absence of a specific keyword prompt. The scene prompt provides more general information about the type of keyword being detected, including different types of inputs such as default wake words, custom wake words, noise, rapid speech or non-target speech (\textless dkws\textgreater,\textless ckws\textgreater,\textless noise\textgreater, \textless rapid\textgreater, \textless non-target\textgreater ). These elements can also appear in combination, for example, \enquote{\textless  noise\textgreater,\textless  dkws\textgreater} represents the detection of default wake words in a noisy environment. Two training samples of PromptKWS with different scene are shown in Table~\ref{tab:tab1}

\begin{table}[htbp]  
\centering  
%\captionof{table}{Two training samples with different type of prompts.}
\caption{Two training samples with different type of prompts.}
\label{tab:tab1} 
\begin{tabular}{|c|c|}  
\hline  
keyword & XIAO KE AI \\ \hline  
scene prompt & \textless noise\textgreater,\textless ckws\textgreater \\ \hline  
keyword prompt & XU XIAO YOU \\ \hline\hline  
keyword & HI Quark \\ \hline  
scene prompt & \textless rapid\textgreater \\ \hline  
keyword prompt & \textless no\_prompt\textgreater \\ \hline  
\end{tabular} 
\end{table}
%\end{center}    

\subsection{Training Criterion}

In the PromptKWS, the loss function is comprised of two parts: the CTC loss from the KWS network, and the cross-entropy loss from the PPN. The CTC loss is used to measure the difference between the predicted and true keyword label sequences, while the cross-entropy loss is used to measure the difference between the predicted and true prompt phrase labels. The joint loss function for the model is defined as: 

\begin{equation}
L=\lambda L_{CTC}+(1-\lambda) L_{PPN}
\label{equation:eq2}
\end{equation}
Here, L represents the total loss function, \(\lambda\) is the weight, and L\textsubscript{CTC} and L\textsubscript{PPN} represent the CTC loss and PPN loss, respectively.

\section{Experiments}

\subsection{Model Configuration}

The baseline model we used is a 5-layer conformer \cite{conformer}, each layer having a dimension of 128 and utilizing multi-head attention with 4 attention heads.  The PPN encoder consists of 3 transformers \cite{att} layers. Each layer includes multi-head attention with 4 heads, feedforward networks with ReLU activation functions, and layer normalization. In the PPN encoder, the dimension of attention was set to 128, and the dimension of the FF layer was set to 512. The PPN decoder consists of 2 transformer layers. 

The model, comprising several components, has a total of 2.7M parameters. These components encompass an encoder with 2.07M parameters (of which 0.31M are parameters of MHCA), a CTC decoder with 0.11M parameters, a PPN encoder with 0.66M parameters, and a PPN decoder with 0.44M parameters. The output units are comprised of 436 context-independent (CI) phones, supplemented by an \enquote{\textless unk\textgreater} symbol and a \enquote{\textless no\_prompt\textgreater} symbol. It should be noted that the PPN encoder is only operational during system initialization, whereas the PPN decoder is exclusively utilized during the training stage. This efficient architecture results in the promptkws model adding a mere 0.97M extra parameters to the baseline model, a quantity that is insignificant when contrasted with the use of a pre-trained BERT model in PromptASR. Therefore, the model is suited for deployment on edge devices.

\subsection{Datasets}

We trained our models on a 5000-hour Mandarin ASR corpus, derived exclusively from speech assistant products. To guarantee randomness, we shuffled the development set, which was sampled from the training dataset. For accurate evaluation of our models' performance, we created a general positive test set D\textsubscript{general}, recorded in an identical environment to the training set. This test set encompassed 100 different keywords, ranging from two to five Chinese characters in length, with 90 samples per keyword. we used a separate audio set of 100 hours, which primarily consisted of background noise, music, wind noise, machine noise and unrelated speech as the negative test set. 
We also constructed a few-shot test set D\textsubscript{few} to evaluate the effectiveness of keyword prompts in low sample scenario. This dataset includes including 0-shot, 1-shot, 5-shot, 10-shot, and 15-shot scenarios, is specifically designed to evaluate scenarios where there are limited training examples for a given keyword. Additionally, we created two special scenario test set D\textsubscript{rapid} and D\textsubscript{noisy} to evaluate the impact of keyword prompts in challenging environments, including high background noise and rapid speech.
To further validate our framework's performance and ensure reproducibility, we also conduct experiments on the public Mandarin benchmark, MobvoiHotwords \cite{mobvoi}. This dataset contains approximately 262 hours of audio featuring two keywords, "Hi Xiaowen" and "Nihao Wenwen", recorded in various real-world noisy environments.

\subsection{Evaluation Metrics}

In order to perform a thorough evaluation of our model's performance, we make use of two key metrics: the receiver operating characteristic (ROC) curve and the accuracy (ACC). For the ROC curves, we generate individual curve for each keyword by utilizing positive samples in conjunction with the 100-hour negative set. The x-axis of the ROC curve represents the false alarm rate per hour (FA), and the y-axis represents the false rejection rate (FRR).

\subsection{Performance of PromptKWS}

First, we compared the accuracy of different KWS models. As shown in Table~\ref{tab:tab2}, our proposed method significantly outperforms the baseline across all three datasets: D\textsubscript{general}, D\textsubscript{rapid}, and D\textsubscript{noisy}. Specifically, our method achieved an accuracy of 92.3\% on D\textsubscript{general}, an improvement of 11.4\% over the baseline. On D\textsubscript{rapid}, our accuracy was 87.9\%, a gain of 13.8\%. Most notably, on D\textsubscript{noisy}, our method boosted accuracy by 15.9\% to 85.4\%. These results indicate that our model demonstrates superior performance in KWS tasks across various scenarios, especially in more challenging fast speech and noisy environments.

\begin{table}[htbp]  
\centering  
\caption{Comparison of ACC for different models.}  
\label{tab:tab2}  
\resizebox{0.45\textwidth}{!}{
\begin{tabular}{ccccc}  
\toprule  
\small\textbf{Method} & \multicolumn{3}{c}{\small\textbf{ACC(\%)}} \\  
\cmidrule(lr){2-4}  
& \small\textbf{D\textsubscript{general}} & \small\textbf{D\textsubscript{rapid}} & \small\textbf{D\textsubscript{noisy}} \\  
\midrule  
\small Baseline & 81.7 & 74.1 & 69.5 \\  
\small + Prompt-acoustic MHCA  & 89.5 & 84.6 & 82.2 \\  
\small + Prompt Phrases Prediction loss & 92.3 & 87.9 & 85.4 \\  
\midrule  
\small\textbf{PromptKWS} & \textbf{92.3} & \textbf{87.9} & \textbf{85.4} \\ 

\bottomrule  
\end{tabular}
}
\end{table}

\begin{figure}[ht]
  \centering
  \includegraphics[width=\linewidth]{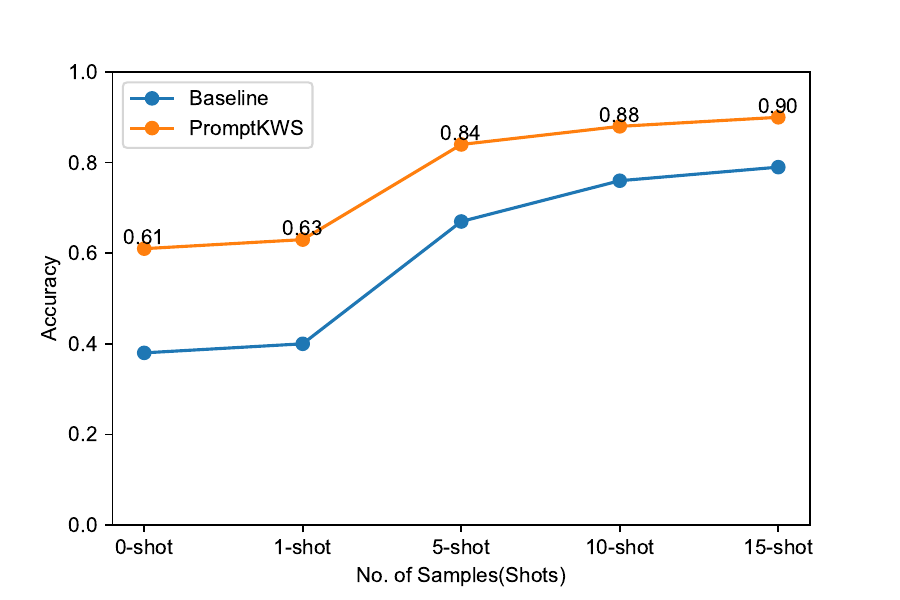}
  \caption{Accuracy for few-shot set.}
  \label{fig:feshot}
\end{figure}

In Figure~\ref{fig:feshot}, we show the performance of  Baseline and PromptKWS in D\textsubscript{few}. From the results, we can observe that the PromptKWS model consistently outperforms the Baseline model across all scenarios. 

Specifically, the accuracy of the Baseline model ranges from 38.1\% to 79.0\% as the number of samples increases from 0 to 15. In contrast, the PromptKWS model achieves a significantly higher accuracy, ranging from 61.0\% to 90.0\%. Notably, even in the 0-shot scenario, where no sample is available, the PromptKWS model still achieves an accuracy of 61.0\%, which is 22.9\% higher than the Baseline model. This demonstrates the superior zero-shot learning capability of the PromptKWS model. As the number of samples increases, the performance of both models improves and the PromptKWS model maintains a clear advantage over the Baseline model, achieving an accuracy of 90.0\% in the 15-shot scenario, which is 11.0\% higher than the Baseline model.
\subsubsection{Effectiveness of Prompt-acoustic MHCA}
By integrating keyword prompt embedding via Prompt-acoustic MHCA into the baseline model, we obtain the Prompt-guided model. As shown in the second row of Table~\ref{tab:tab2}, when compared to the baseline model, the Prompt-guided model exhibits a significant improvement in performance across all test sets, particularly in more challenging scenarios. Specifically, the accuracy rates in noisy and rapid speech scenarios increased by 10.5\% and 12.7\%, respectively. This demonstrates that keyword prompting guided by Prompt-acoustic MHCA is an exceptionally effective technique for open-vocabulary KWS. 
\subsubsection{Effectiveness of PPN}
As shown in the third row of Table~\ref{tab:tab2}, the PPN introduced in Sec 2.2 can further improve performance on the guided model.  When the PPN loss is added to the model, there is an average accuracy improvement of 3.1\% across all test sets. This result strongly confirms the crucial role played by our proposed PPN loss in extracting more discriminative embeddings.

\begin{figure}[ht]
  \centering
  \includegraphics[width=\linewidth]{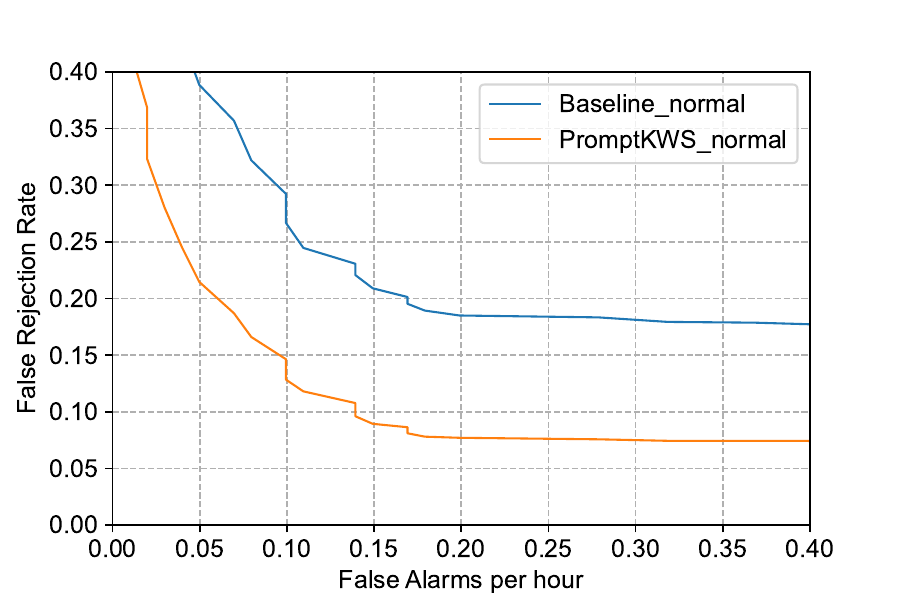}
  \caption{ROC curves for general speech scenario.}
  \label{fig:normal}
\end{figure}
\begin{figure}[ht]
  \centering
  \includegraphics[width=\linewidth]{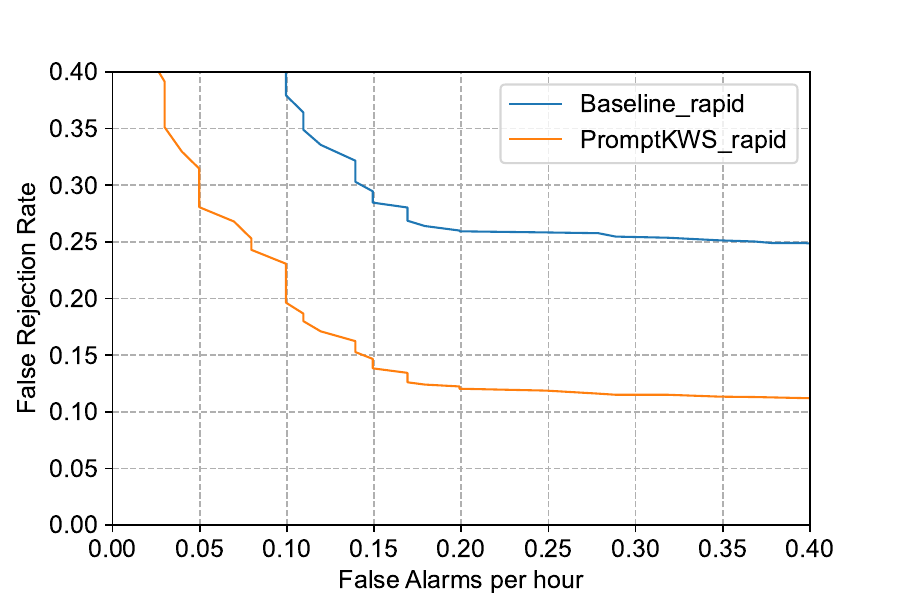}
  \caption{ROC curves for rapid speech scenario.}
  \label{fig:rapid}
\end{figure}
\begin{figure}[ht]
  \centering
  \includegraphics[width=\linewidth]{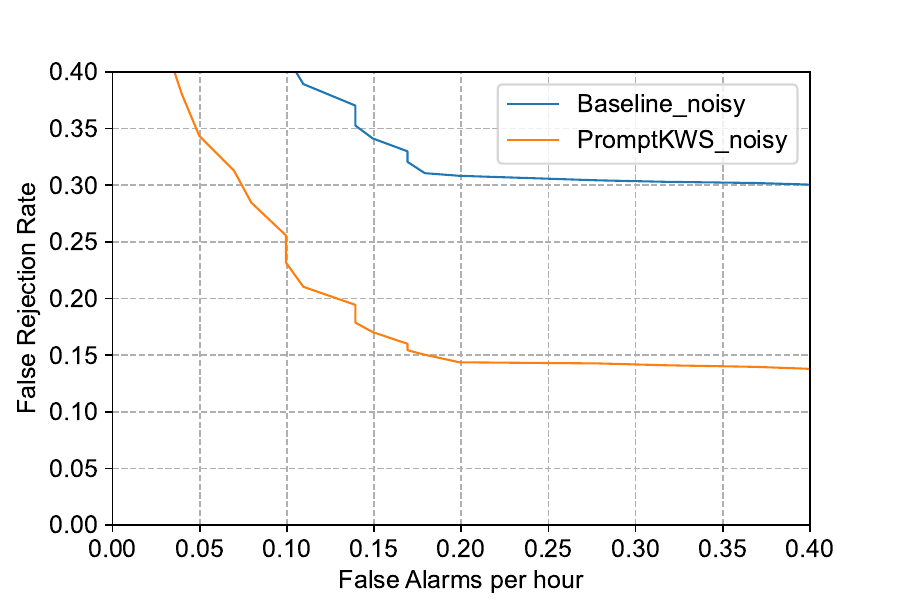}
  \caption{ROC curves for noisy speech scenario.}
  \label{fig:noisy}
\end{figure}

\subsection{Performance Analysis via ROC Curves}

For a comprehensive analysis, we constructed the Receiver Operating Characteristic (ROC) curve and comparing the performance of the baseline model with that of our proposed PromptKWWS framework. Figure~\ref{fig:normal},~\ref{fig:rapid},~\ref{fig:noisy} present ROC curves  under different speech scenarios: general, rapid, and noisy. 

In the general scenario, the PromptKWS model significantly outperforms the Baseline model, achieving a 11\% improvement in the wake-up rate at 0.2 FA. This is evident as the FRR for PromptKWS is 0.077 compared to 0.187 for the Baseline model at 0.2 FA. In the rapid speech scenario, our method consistently outperforms the baseline. The FRR of the PromptKWS model is only 0.13 at 0.2 FA, compared to the Baseline model's FRR of 0.26. This indicates an improved performance by approximately 13\%, showing that our PromptKWS model is highly effective in fast speech conditions. In noisy scenarios, the presence of noise poses a challenge for KWS systems, leading to a notable increase in the FFR under the same FA conditions. Despite the challenging conditions, the superiority of the PromptKWS model is even more pronounced. The FRR of the PromptKWS model at zero false alarms is 0.146, while the Baseline model has a FRR of 0.309. This represents a 16\% improvement, highlighting the robustness of our PromptKWS model in noisy conditions.

\section{Conclusions}

In this paper, we present PromptKWS, a novel approach for open-vocabulary KWS. Our model employs a PPN moudle and a Prompt-guided KWS encoder. This encoder is based on the Prompt-acoustic MHCA, which allows the model to dynamically adapt to keyword according to the keyword prompt. The PromptKWS model boasts a compact size of 2.7M, making it ideal for deployment on edge devices.To demonstrate the efficacy of our proposed PromptKWS framework, we conducted a series of experiments. These experiments assessed the model's performance in various challenging scenarios, such as low-sample, rapid speech and high-noise environments. Our results show that PromptKWS significantly outperforms the baseline in a wide range of scenarios, highlighting its superiority and adaptability in real-world applications.

\bibliographystyle{IEEEtran}
\bibliography{mybib}

\end{document}